# Toward Natural Gesture/Speech Control of a Large Display

Sanshzar Kettebekov, Rajeev Sharma

Department of Computer Science and Engineering
Pennsylvania State University
220 Pond Laboratory, University Park PA 16802
`sanshzar@psu.edu, rsharma@cse.psu.edu`

**Abstract.** In recent years because of the advances in computer vision research, free hand gestures have been explored as means of human-computer interaction (HCI). Together with improved speech processing technology it is an important step toward natural multimodal HCI. However, inclusion of non-predefined continuous gestures into a multimodal framework is a challenging problem. In this paper, we propose a structured approach for studying patterns of multimodal language in the context of a 2D-display control. We consider systematic analysis of gestures from observable kinematical primitives to their semantics as pertinent to a linguistic structure. Proposed semantic classification of co-verbal gestures distinguishes six categories based on their spatio-temporal deixis. We discuss evolution of a computational framework for gesture and speech integration which was used to develop an interactive testbed (iMAP). The testbed enabled elicitation of adequate, non-sequential, multimodal patterns in a narrative mode of HCI. Conducted user studies illustrate significance of accounting for the temporal alignment of gesture and speech parts in semantic mapping. Furthermore, co-occurrence analysis of gesture/speech production suggests syntactic organization of gestures at the lexical level.

## 1. Introduction

Psycholinguistic studies, e.g., McNeill [1] describes gestures as a critical link between our conceptualizing capacities and our linguistic abilities. People use gestures to convey what cannot always be expressed using speech only. These gestures are sometimes considered as medium for expressing thoughts and ways of understanding events of the world. Kendon [2] distinguishes *autonomous gestures* from *gesticulation*, gestures that occur in association with speech. In fact when combined with speech they were found to be efficient means of communication for coping with the complexity of the visual space, e.g., pen-voice studies [3]. Motivated by this, there has been a considerable interest in integrating gestures and speech for natural human-computer interaction .
 To date, speech and gesture recognition have been studied extensively but most of the attempts at combining speech and gesture in an interface were in the form of a predefined controlled syntax such as "*put <point> that <point> there*", e.g., [4]. In a number of recent applications user was confined to the predefined gestures (signs) for spatial browsing and information querying, e.g. [5]. In both cases the intent of making interaction natural was defeated.

Part of the reason for the slow progress in multimodal HCI is the lack of available sensing technology that would allow sensing of natural behavior on the part of the user. Most of the researchers have used some device (such as an instrumented glove or a pen) for incorporating gestures into the interface, e.g., [6]. This leads to constrained interaction that result in unnatural multimodal patterns. For instance, in the pen-voice studies [5] the order and structure of incoming linguistic information significantly deviated from the ones found in canonical spoken English. However, the availability of abundant processing power has contributed to making computer vision based continuous gesture recognition robust enough to allow the inclusion of free hand gestures in a multimodal interface [7, 8, 9].

An important feature of a natural interface would be the absence of predefined speech and gesture commands. The resulting multimodal "language" thus would have to be interpreted by a computer. While some progress has been made in the natural language processing of speech, there has been very little progress in the understanding of multimodal human-computer interaction [10].

Although, most of gestures are closely linked to speech, they still present meaning in a fundamentally different form from speech . Studies in human-to-human communication, psycholinguistics, and etc. have already generated significant body of research. However, they usually consider a different granularity of the problem outside the system design rationale. When designing a computer interface even when it incorporates reasonably natural means of interaction, we have to consider artificially imposed paradigms and constraints of the created environment. Therefore exclusively human studies cannot be automatically transferred over to HCI. Hence, a major problem in multimodal HCI research is the availability of valid data that would allow relevant studies of different issues in multimodality. This elaborates into a "chicken-and-egg" problem.

To address this issue we develop a computational methodology for the acquisition of non-predefined gestures. Building such a system involves challenges that range from low-level signal processing to high-level interpretation. In this paper we take the following approach to address this problem. We use studies in an analogous domain; that of TV weather broadcast [7] for bootstrapping an experimental testbed, *i*MAP. At this level, it utilizes an interactive campus map on a large display that allows a variety of spatial planning tasks using spoken words and free hand gestures. Then we address critical issues of introducing non-predefined gesticulation into HCI by conducting user studies.

The rest of the paper is organized as follows. In Section 2 we review suitable gestures and their features for the display control problem. Subsequently (in Section 3) we present structural components of gestures as pertinent to a linguistic structure. A semantic classification of co-verbal gestures is also presented in this section. Section 4 introduces a computational framework of the gesture/speech testbed and describes results of continuous gesture recognition. Section 5 presents results of *iMAP* user studies.

## 2. Introducing Gesticulation into HCI

Kendon [2] distinguishes *autonomous gestures* from *gesticulation*, gestures that occur in association with speech. McNeill [1] classifies major types of gestures by their relationship to the semantic content of speech, in particular a person's point of view towards it. He defines *deictic* gestures that indicate objects and events in the concrete world when their actual presence might be substituted for some metaphoric picture in one's mind. *Iconic* gestures were those that refer to the same event as speech but present a somewhat different aspect of it. For example: "and he bends <*extending bent arm in the elbow*> …". The other types are metaphoric and beat gestures, which usually cover abstract ideas and events. See [1] for a detailed review. The last two gesture classes have currently a limited use for HCI.

As for HCI, deictic gestures seem to be the most relevant. This term is used in reference to the gestures that draw attention to a physical point or area in course of a conversation [1]. These gestures, mostly limited to the pointing by definition, were found to be co-verbal [1, 11]. Another type of gestures that may be useful is sometimes referred to as manipulative or act gestures. These serve to manipulate objects in virtual environment e.g., [12]. They usually characterized by their iconicity to a desired movement, e.g., rotation. Later in the section 3.3 we will present classes of co-verbal gestures based on their deixis that were found useful during narrative display control.

To date, very limited research has been done on how and what we actually perceive in gestures. In face-to-face communication some of the gestures may be perceived on the periphery of the listener's visual field [2, 13]. In that case, the gestures were not visually attended and their meaning may be mainly conveyed by their kinematical patterns. Supporting arguments exist in recent eye tracking studies of anthropomorphic agents while gesturing [14]. They revealed that attention of the listener shifts to the highly informative one-handed gesture when it had changes in a motion pattern, i.e., a slower stroke.

In addition, a shape of the hand while gesturing can be affected by an uncountable number of cultural/personal factors. Therefore, considering manipulation on a large display our approach concentrates on the kinematical features of the hand motion rather then hand signs. Section 4.2 will present review of continuous recognition of gesture primitives[1] based on their kinematical characteristics.

Even though gestures are closely linked to speech, McNeill [1] showed that natural gestures do not have a one-to-one mapping of form to meaning and their meaning is context-sensitive. In the next section we will attempt to clarify nature of co-verbal gestures as pertinent to a linguistic structure. Our approach is dictated by a design rational that begins from acquisition of kinematical gesture primitives to extraction of their meaning.

---

[1] In the gesture literature it is also referred as a gesture form or a phoneme (sign language).

## 3. Understanding of Gestures

Armstrong [15] views gestures as a neuromuscular activity that ranges from spontaneously communicative to more conventional gestures that poses linguistic properties when presented as conventualized signs. The formalist approach assumes that speech can be segmented into a linear stream of phonemes. Semantic phonology introduced by [16] is a reasonable approach for representing structural organization of gestures and similar to Langacker's notion of cognitive grammars; it reunites phonology with the cognitive, perceptual, and motoric capacities of human.

Since gestures realized in spatial medium through optical signal we would expect that gesture phonemes defined in the spatio-temporal domain progress to form morphemes (morphology), the words in turn have different classes that are used to form phrase structures (syntax), and finally those yield meaning (semantics). A semantic phonology ties the last step to the first, making seamless connection throughout the structure. Cognitive grammar [17] claims that a linguistic system comprises just three kinds of structures: phonological, semantic, and symbolic (residing in the relationship between a semantic and a phonological structure). In the following subsections we will address those structures from a computational rationale.

### 3.1. Multidimensionality of Meaning

At the phonological level, a continuous hand gesture consists of a series of qualitatively different kinematical phases such as movement to a position, hold, and transitional movement [2]. Kendon [18] organized these into a hierarchical structure. He proposed a notion of gestural unit (*phrase*) that starts at the moment when a limb is lifted away from the body and ends when the limb moves back to the resting position. This unit consists of a series of gesture *morphemes* [2], which in turn consists of a sequence of *phonemes*: *preparation—stroke—hold—retraction* (Fig. 1). The *stroke* is distinguished by a peaking effort. When multiple gesture morphemes are present, the *retraction* can be either very short or completely eliminated. Kendon concluded, a morpheme is the manifestation of a so-called idea unit. [18] found that different levels of movement hierarchy are functionally distinct in that they synchronize with different levels of prosodic structuring of the discourse in speech. These results will be elaborated in the next subsection.

At the morphological level, McNeill [1] argued that gestures occurring in face-to-face communication have global–synthetic structure of meaning. The meaning of a part of a gesture is determined by the whole resulting gesture (e.g., global), where different segments are synthesized into a single gesture (e.g., synthetic). Then if a gesture consists of several strokes, the overall meaning of each stroke is determined by the meaning of the compounded gesture.

Stokoe [19] proposed to treat sublexical components of visual gestures (signs) at as motion, irrespective of what is moving; the hand configuration as it was at rest; and the location where the activity occurs. Figure 1 shows compositional semantics of gesture strokes applicable in HCI; we define it as being distributed among, a spoken clause, and a physical (spatial) context. In addition the world-level units of the language of a visual gesture can be viewed as a marriage of a noun and a verb; an action and something acting or acted upon. For simplicity (not incompleteness): such a system encapsulates both words and sentences, for the words can be symbolically, SVO (subject-verb-object) [19]. We will consider these notions in classification of the co-verbal gestures in Section 3.3.

---

[2] Kendon refers to this unit as gesture phrase (G-phrase)

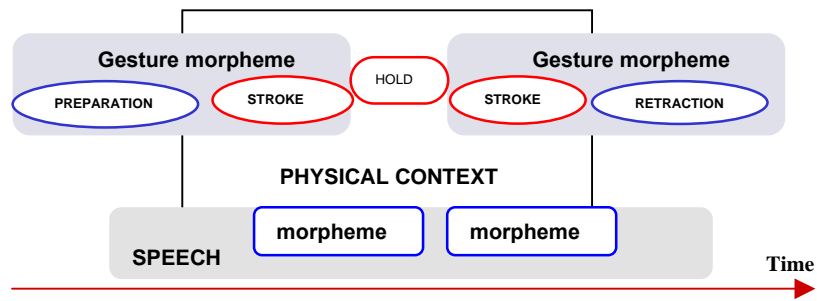

**Fig. 1.** Structural semantics of co-verbal gestures. Strokes are meaningful parts of gesture morphemes as defined in the context. Strokes and holds are temporally aligned with the relevant spoken words.

### 3.2. When Meaning Is Formed

The issue of how gestures and speech relate in time is critical for understanding the system that includes gesture and speech as part of a multimodal expression. McNeill [1] defines a semantic synchrony rule which states that speech and gestures cover the same idea unit supplying complementary information when they occur synchronously.

The current state of HCI research provides partial evidence to this proposition. Our previous studies of weather narration [7] revealed that approximately 85% of the time when any meaningful gestures are made, it is accompanied by a spoken keyword mostly temporally aligned during and after the gesture. Similar findings were shown by Oviatt et al. in the pen-voice studies [11].

At the phonological level, Kita et al. [20] assert that a pre-stroke hold is more likely to co-occur with discourse connectors such as pronouns, relative pronouns, subordinating temporal adverbials (e.g., when, while, as) compared to a post-stroke hold. In an utterance, the discourse cohesion is established in the initial part, and new information is presented in the final part. It is thought that a pre-stroke hold is a period in which gesture waits for speech to establish cohesion so that the stroke co-occurs with the co-expressive portion of the speech. [20] found that a stroke with repetitive motion is less likely to follow post-stroke hold. It was suggested that a post-stroke hold was a way to temporally extend a single movement stroke so that the stroke and post-stroke hold together will synchronize with the co-expressive portion of the speech. Despite the fundamental structural differences of gestures and speech the co-occurrence patterns imply syntactic regularities during message production in face-to-face communication.

### 3.3. Semantic Classification of Gestures

After extensive studies of the weather channel narration [7, 8] and pilot studies on the computerized map we separated two main semantic categories. *Transitive* deixis embraces all the gesture acts that reference concrete objects (e.g., buildings) in the context where no physical change of location was specified as. In the literature these gestures are also referred as the *deictic* gestures [1]. In HCI studies those often were found to be supplemented by a spoken deictic marker (this, those, here, etc.) [11]. Unlike traditionally accepted gesture classification [1] where only the pointing gesture is referred as deictic, we define three subcategories of deictic strokes independently of their form (Figure 2).

Categorization of *transitive* deixis is inferred from the complementary verbal information. They are defined as follows: *nominal, spatial,* and *iconic. Nominal* refers to object selection made through reference of the object itself by an assisting noun or pronoun. *Spatial* deixis is made through an area that includes the reference and usually complemented by adverbials "here", "there", "below", and etc. The *iconic* category is a hybrid of *nominal* and *spatial* deixis. The gesture in this case not only refers to the object but also specifies its attributes in a spatial continuum, e.g., shape. For example, "**this** road <*contour* stroke along>".

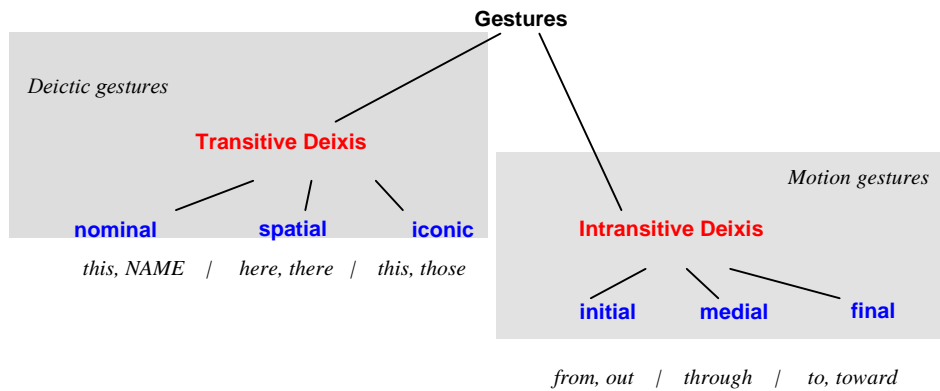

**Fig. 2.** Semantic classification of gestures and examples of the associated keywords

*Intransitive* deixis is defined as a gestural act causing spatial displacement. This category is analogous to the spoken motion verb (e.g., move, enter). Unlike *transitive* deixis instead of an object, it requires a spatio-temporal continuum, i.e. direction, as a reference to complete the meaning. Analogous to Talmy's classification [21] of intransitive motion verbs, we distinguish three classes: *initial, medial,* and *final*. These are characterized by a restructuring of space, i.e. spatial displacement, induced by reference-location. These gestures are expected to co-occur with the dynamic spatial propositions (from, through, toward, etc.). For example, in "go **through** <stroke *through* > the building" the stroke corresponds to the *medial* motion gesture.

## 4. Addressing the "Chicken -and -Egg" Problem

Realization of the issues discussed so far in a computation framework, derived from audio and video signals, is a difficult and challenging task. The integration of gesture and speech can occur in three distinct levels - data, feature, or decision level [22]. Data fusion is the lowest level of fusion. It involves integration of raw observations and can occur only in the case when the observations are of the same type which is not common for multiple modalities. Feature level fusion techniques are concerned with integration of features from individual modalities into more complex multimodal features and decisions. These frequently employ structures known as probabilistic networks, such as artificial neural networks and hidden Markov models (HMMs).

Decision fusion is the most robust and resistant to individual sensor failure. It has a low data bandwidth and is generally less computationally expensive than feature fusion. However, a disadvantage of decision level fusion is that it potentially cannot recover from loss of information that occurs at lower levels of data analysis, and thus does not exploit the correlation between the modality streams at the lower integration levels. Frame-based decision level fusion have been commonly used ever since Bolt's early "Put- that-there" [4].

Both decision and feature level require *natural* multimodal data, e.g., for utilizing statistical techniques. This controversy leads to a "chicken-and-egg" problem. One of the solutions is to use Wizard-of-Oz style of experiments [23]. Where, the role of a hypothetical multimodal computer is played by a human "behind the scene". However, this method does not guarantee a timely and accurate system response which is desirable for eliciting adequate user interaction. Therefore, in the next subsection we present a gesture/speech testbed enabled with gesture recognition (primitives) and timely feedback.

### 4. 1. Computational Framework: *i*MAP

Figure 3 presents an overview of the conceptual design of the multimodal HCI system composed of three basic levels. At the topmost level, the inputs to the *i***MAP** are processed by low-level vision based segmentation, feature extraction and tracking algorithms [24], and speech processing algorithms. The output of this level goes to two independent recognition processes for speech and gesture. The gesture recognition framework has been inspired and influenced by a previous work for continuous gesture recognition in the context of a weather map [7]. In order to achieve an adequate perceptual user interface *i***MAP** utilized both audio and visual feedback.

The position of the cursor is controlled by the hand by means of a 2D vision-based tracking process, for details see [24]. It enabled rather steady and near real time positioning of the cursor with up to 110° (visual angle) sec$^{-1}$. The visual (building name) and audio (click) feedback was enabled by the recognition of a basic *point* and consecutive *contour* gesture. When the pointing gesture was recognized the cursor shape is changed from "🖐" to "☚" (Fig. 4.b).

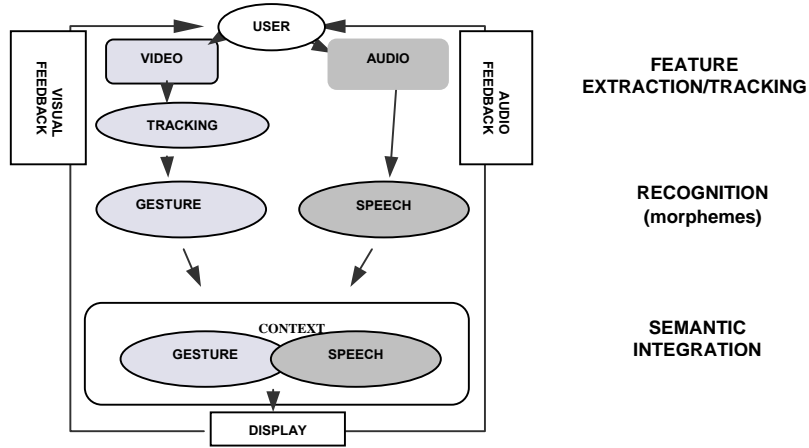

**Fig. 3.** Framework diagram for gesture/speech testbed.

### 4.2. Continuous Gesture Recognition

We sought a solution to bootstrapping continuous gesture recognition (phonemes) through the use of an analogous domain, which does not require predefinition of gestures. We refer to it as the *weather domain*. The weather domain is derived from the weather channel on TV that shows a person gesticulating in front of the weather map while narrating the weather conditions (Fig. 4.a).

Here, the gestures embedded in a narration are used to draw the attention of the viewer to specific portions of the map. Similar set of gestures can also be used for the display control problem. The natural gestures in the weather domain were analyzed with the goal of applying the same recognition techniques to the design of a gesture recognition system for *i*MAP.

To model the gestures both spatial and temporal characteristics of the hand gesture need to be considered. In [7] the natural gestures were considered to be adequately characterized by 2D positional and time differential parameters. The meaningful gestures are mostly made with the hand further away from the body. Little or no movement characterizes other gestures that do not convey any information with the hands placed close to the body e.g., beat gestures.

Analysis of the weather narration [7] made it possible to discern three main strokes of gestures that are meaningful in spite of a large degree of variability. Those were termed as *pointing*, *circle,* and *contour* primitives (Fig. 5).

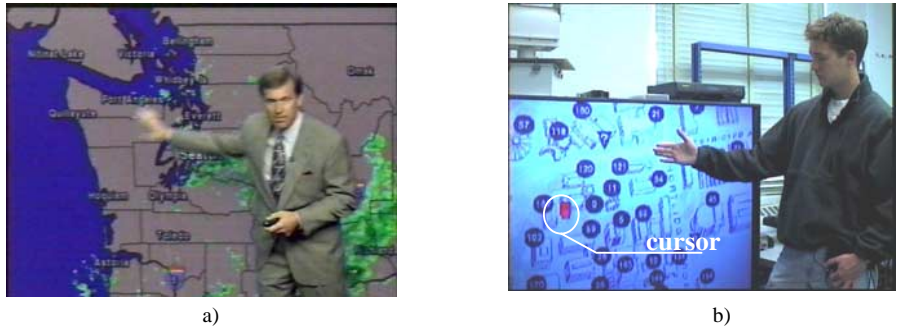

**Fig. 4.** Two analogous domains for the study of natural gestures: a) The weather narration and b) The computerized map interface (*i*MAP)

Since compound gesture can be formed by a combination of the strokes of one or more of the aforementioned three primitive categories. The preliminary experiments led to the choice the following HMMs which were constrained as shown in the morphological network (Fig. 5) [7]. Left-to-right causal models were defined, with three states for each of the *preparation, retraction*, and *point* HMMs and four states for each of the *contour* and *rest* HMMs.

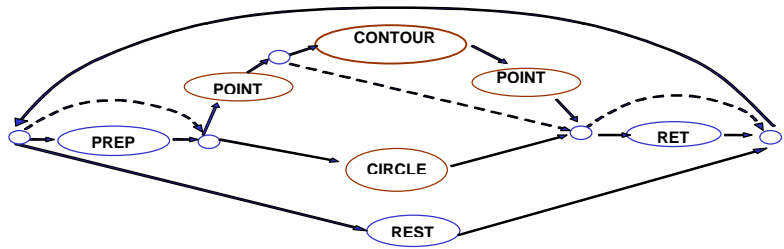

**Fig. 5.** Morphological network of gesture primitives. Dashed lines imply an alternative transition graph to a next phoneme. Meaningful strokes of gestures are identified as gesture primitives: *contour*, *circle*, and *point*.

### 4.3. Results in natural gesture recognition

The results from extensive experiments conducted in both the weather domain and the ***i*MAP** framework indicated that it is possible to recognize natural gestures continuously with reasonably good rates, 78.1 and 79.6% respectfully [7, 8]. Method of the preliminary bootstrapping in the analogous domain proved to be an effective solution.

Consequent application of speech/gesture co-occurrence model in the weather domain through the associated keyword spotting (noun, pronoun, and spatial adverbials) significantly improved gesture recognition. A correct rate of 84.5% was attained [7]. The next section presents a series of user studies that contributes to formalization of a framework for meaningful gesture processing at the higher levels.

## 5. Toward Eliciting Syntactic Patterns

The goal of the following study is to elicit empiric clarification on the underlying semantics of natural gestures and syntactic patterns for the display control problem. We adopt classes of gestures from [14, 5]: *point*, *contour*, and *circle*. We consider a series of spatial tasks in the context of a campus map. Those included combination of object selection/localization tasks (e.g., "What is the nearest parking lot?") and their narrative displacement to other locations (e.g., "How would you take your car to another location?"). The presented a set of spatial tasks did not require map manipulation, e.g., zooming, rotation, scrolling. See [25] for detailed review.

**5.1. User Study Results: Multimodal Patterns**

This section presents a summary *i*MAP user study results, c.f. [25], that was conducted with 7 native English speakers. Analysis of a total of 332 gesture utterances revealed that 93.7% of time the adopted gesture primitives were temporally aligned with the semantically associated nouns, pronouns, spatial adverbials, and spatial prepositions. Unlike previous attempts e.g., [11, 8] it permitted us to take a first step in the investigation of the nature of *natural* gesture-speech interaction in HCI.

*Transitive* deixis was primarily used for the selection/localization of the objects on the map. *Intransitive* deixis (gestures that were meant to cause spatial displacement) were mostly found to express the direction of motion.

The original classification of gesture primitives was supported by stronger intrinsic separation for *intransitive* deixis compared to the *transitive* category. This is due to the kinematical properties of the *point* and *contour* gestures, which are more likely to express the direction of the motion. The utterance analysis revealed presence of motion complexes that are compounded from the classes of *intransitive* deixis. *Point* was found to be the only primitive to express *initial movement* deixis which is followed by either *point* or *contour* strokes depending on the type of motion presented, i.e. implicit or explicit respectively. The explicit type was mostly encountered if precise path need to be specified.

Due to associative bias of the *contour* primitive with the motion, the temporal co-occurrence analysis was found effective in establishing pattern for the following primitive. If an *initial* spatial preposition co-occurred with the pre-stroke hold of the *contour* it was followed by *point* primitive to establish destination point of motion. In, general this finding agrees with Kita [20] that pre-stroke hold is indicative of the discourse function.

The results indicate that the proposed method of semantic classification together with co-occurrence analysis significantly improve disambiguation of the gesture form to meaning mapping. This study, however, did not consider perceptual factors in the visual context. Studies by De Angeli et al. [3] showed that salience and topological properties of the visual scene, i.e., continuity and proximity affect the form of gesture, similar to the *iconic* deixis in our classification. Inclusion of this notion should contribute to the disambiguation in the *transitive* deixis category.

**5.2. Syntactic Patterns**

In human-to-human studies McNeill argues that gestures do not combine to form larger, hierarchical structures [1]. He asserts that most gestures are one to a clause, but when there are successive gestures within a clause, each corresponds to an idea unit in and of itself. In contrast to those findings the current results suggest that the gesture primitives (phonemes), not gestures as defined by McNeill, could be combined into a syntactic binding. The primitives were treated as self-contained parts with the meaning partially supplemented through the spatial context and the spoken counterpart. From the established patterns we can combine gesture phonemes that were mapped into the *intransitive* deixis, into the motion complexes (Fig. 6).

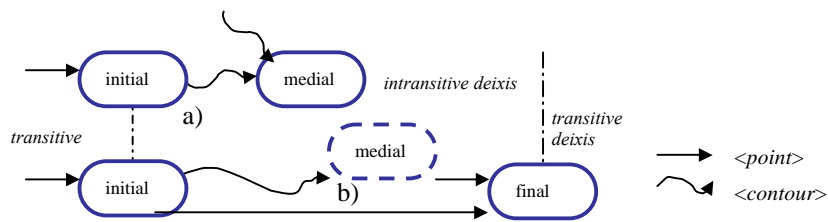

**Fig. 6.** Motion complexes of gestures and their relationships to the deixis categories: a) *Intransitive* complex*: initial* or *medial* gestures that precede or occurs synchronously with the speech do not result in the *final* stroke; b) *Transitive* complex*: medial* class that has a pre-stroke hold and eventually is followed by the *transitive* gesture (*point*)

Since *transitive* deixis is inherently a part of the motion complex, which is likely to have a compounded structure, we distinguish two syntactic structures: *transitive* and *intransitive* motion complexes. In case of *intransitive* complex a gesture stroke conveying movement, e.g., *contour*, classified as *medial* is not usually followed by another stroke if the stroke precedes or occurs synchronously with a spoken keyword. E.g., "take this <*point*> car <*contour*> **out** of the lot". *Transitive* motion complex is characterized by a *medial* pre-stroke hold often followed by the final *pointing*. This formulation of spatio-temporal semantics of gesture links individual stroke and sequence of strokes used in "go form here to here" constructions. The emergence of the above patter is similar to SVO (subject-verb-object) structure proposed by Stokoe [16]. This system encapsulates both word-level and sentence-level structures.

This formulation became possible when our method considered semantic categorization of spatio-temporal properties of gestures (Fig. 2). Unlike McNeill's classification of gestures [1] it distinguishes between morphological (primitives) and functional characteristics (spatio-temporal context). Consideration of spatio-temporal context (semantic classification and co-occurrence models) at the phonological level enabled us to distinguish two levels of language structure in the co-verbal gestures: morphological and lexical. These findings are applicable in the context of spatial manipulation on a large display and may not have analogies in the human-to-human communication.

## 6. Conclusions

There is a great potential for exploiting the interactions between speech and gesture to improve HCI. A key problem in carrying out experiments for gesture speech recognition is the absence of truly natural data in the context of HCI. Our approach was to use the study of gesture and speech in the analogous domain of weather narration to bootstrap an experimental testbed (*i*__MAP__) achieving reasonably good recognition rates. This in turn provided the testbed for relevant user studies for understanding the gestures in the context of HCI. *i*__MAP__ user studies indicated highly multimodal interaction. Results of the current study allow us to sketch mechanics of interpretation process when dealing with non-predefined gestures. Proposed semantic classification together with gesture speech co-occurrence allowed us to draw causal links of mapping gesture forms to their meaning. Current methodology of combining phonological level and spatio-temporal context (semantic classification and co-occurrence models) enabled us to distinguish morphological and lexical syntactically driven levels of gestural linguistic structure. In general, such systematic approach may provide a foundation for effective framework for multimodal integration in HCI.

**Acknowledgment**

The financial support of the National Science Foundation (CAREER Grant IIS-00-81935 and Grant IRI-96-34618) and U. S. Army Research Laboratory (under Cooperative Agreement No. DAAL01-96-2-0003) is gratefully acknowledged.